\documentclass{article}

\usepackage{microtype}
\usepackage{graphicx}
\usepackage{subfigure}
\usepackage{booktabs} 
\usepackage{CJKutf8}

\usepackage{hyperref}

\usepackage[accepted]{icml2024}

\usepackage{amsmath}
\usepackage{amssymb}
\usepackage{mathtools}
\usepackage{amsthm}
\usepackage{multirow}
\usepackage{diagbox}

\usepackage[capitalize,noabbrev]{cleveref}

\theoremstyle{plain}

\theoremstyle{definition}

\theoremstyle{remark}

\usepackage[textsize=tiny]{todonotes}

\icmltitlerunning{Large Language Models Lack Understanding of Character Composition of Words}

\begin{document}

\twocolumn[
\icmltitle{Large Language Models Lack\\ Understanding of Character Composition of Words}




\begin{icmlauthorlist}
\icmlauthor{Andrew Shin}{yyy}
\icmlauthor{Kunitake Kaneko}{yyy}
\end{icmlauthorlist}

\icmlaffiliation{yyy}{Faculty of Science and Technology, Keio University, Kanagawa, Japan}

\icmlcorrespondingauthor{Andrew Shin}{shin@inl.ics.keio.ac.jp}

\icmlkeywords{Machine Learning, ICML}

\vskip 0.3in
]



\printAffiliationsAndNotice{}  

\begin{abstract}
Large language models (LLMs) have demonstrated remarkable performances on a wide range of natural language tasks. Yet, LLMs' successes have been largely restricted to tasks concerning words, sentences, or documents, and it remains questionable how much they understand the minimal units of text, namely characters. In this paper, we examine contemporary LLMs regarding their ability to understand character composition of words, and show that most of them fail to reliably carry out even the simple tasks that can be handled by humans with perfection. We analyze their behaviors with comparison to token level performances, and discuss the potential directions for future research.
\end{abstract}

\section{Introduction}
\label{intro}
Large language models (LLMs) \cite{Achiam2023GPT4TR,Chowdhery2022PaLMSL,Touvron2023Llama2O,Reid2024Gemini1U,OpenAI2022,Jiang2023Mistral7} have exhibited outstanding performance across a diverse array of natural language tasks. It has largely outperformed pre-LLM approaches on benchmark tasks, such as GLUE \cite{Wang2018GLUEAM} and SuperGLUE \cite{Wang2019SuperGLUEAS}, often surpassing humans on a number of tasks \cite{Chowdhery2022PaLMSL}. It is noteworthy that most of the tasks upon which LLMs have been tested revolve around words, sentences, or passages, but hardly involve character-level understanding. Intuitively, character-level tasks should be much easier to tackle, as they rarely deal with complex semantics, grammatical structures, or background knowledge, while only requiring highly elementary understanding of characters and, depending on the task, simple counting. Indeed, humans are able to perform basic character-level tasks very easily as we will see in Sec~\ref{sec:results}. It has also been known that LLMs hardly make spelling errors and can be used for spelling correction of human-written passages \cite{Whittaker2024LargeLM}. Surprisingly, however, our examination shows that LLMs struggle with very simple tasks involving character composition, severely underperforming humans, making a striking contrast with their performance on more complex tasks at token level.

Humans are able to instantly recognize which characters constitute a given word. However, large language models, most of which are trained at token-level, struggle to grasp the nuances of character composition within words. This difficulty arises from the fact that LLMs primarily learn at the token level, where words are treated as indivisible units separated by spaces or punctuation marks. Consequently, LLMs lack the fine-grained understanding of character-level relationships and morphology that humans possess. Understanding character composition is crucial for various linguistic tasks, including morphological analysis, semantic interpretation, and language generation. As such, addressing the challenge of character composition is essential for enhancing the reliability of LLMs across a diverse range of languages and writing systems.

In this paper, we examine LLMs with a number of simple tasks designed to test the understanding of character composition. None of the tasks requires any advanced knowledge of grammar or semantics, and can be easily tackled with elementary understanding of characters. Yet, our results show a surprisingly poor performance, suggesting that there may be a fundamental drawback with regards to how LLMs are trained and how they perceive the language. We compare LLMs' performances at character level tasks with those at token level tasks of the same types, and investigate the implications of the large discrepancies. We further discuss potential future research directions to enhance LLMs' understanding of character composition, such as incorporating character embedding and visual features into language representation of LLMs.

\section{Related Works}
Although a majority of language models have relied on token-level embeddings, there have been a number of notable endeavors to incorporate character composition or sub-word tokenization into language models, some of which have demonstrated improved performance on relevant tasks. \cite{Kim2015CharacterAwareNL} introduced character-aware neural language models, which utilize character-level embeddings alongside word embeddings to capture morphological and orthographic features of words. Similarly, \cite{wieting-etal-2016-charagram} proposed Charagram, a character-level language model that generates word representations based on character n-grams, enabling better handling of out-of-vocabulary words. \cite{Bojanowski2016EnrichingWV} presented FastText, a fast and efficient word embedding technique that leverages sub-word information to enhance word representations, particularly for morphologically rich languages. While these approaches demonstrate the effectiveness of integrating character information into language models, paving the way for improved performance in various natural language processing tasks, they have mostly been tested on natural language generation tasks, such as Penn Treebank \cite{Marcus1993BuildingAL}, and have not explicitly been tested for understanding of character composition.

Subsequent works in language modeling have further explored the integration of character-level information. For instance, \cite{Peters2018DeepCW} introduced deep contextualized word representations (ELMo), which enhance word embeddings by considering the internal structure of words through character-level convolutions. This method significantly improved the performance of various NLP tasks by capturing complex word morphologies. \cite{Akbik2019FLAIRAE} proposed Flair embeddings, which combine character-level embeddings with contextual string embeddings to provide a more comprehensive representation of words in their context. \cite{clark2020electra} introduced ELECTRA, a pre-training method that includes a discriminative component to identify corruptions at the token level, which indirectly benefits from finer-grained text representations. For most of these works, however, the primary focus has been on token-level tasks rather than specifically addressing character composition understanding. 

With regards to more recent LLMs, there have been a number of works that highlight their downsides from various angles. For example, \cite{Qian2022LimitationsOL} claims that LLMs struggle with arithmetic and symbolic manipulations, while \cite{Lee2024LanguageMD} shows that LLMs fail to learn physical manifestation of language, such as the visuals and sounds of the language. \cite{Truong2023LanguageMA} also shows that LLMs' performances degrade when negation is involved. With regards to the character composition, there have been a few attempts to benchmark the performances of LLMs \cite{Srivastava2022BeyondTI,Efrat2022LMentryAL}, although the scope of evaluating character composition was highly restricted, with stronger emphasis on evaluation of word-level understanding.

\section{Experiments}\label{sec:experiment}
\subsection{Setting}
We perform simple tasks that are designed to assess the LLM's understanding of character composition of words. Nearly all tasks are simple and straightforward with hardly any component for complexity or confusion. It would be fair to state that even humans with very little educational background of up to elementary school can solve most of these tasks without difficulty.

\textbf{Word retrieval}: We provide the LLM with input text and ask it to retrieve all words containing a certain character. For example, ``Find all words that contain the character \textit{h} in the following text: \textit{She is home.}" should output ``\textit{She}" and ``\textit{home}". The task may be examined in variations by specifying the position or the number of occurrences of the characters within a word.


\textbf{Character insertion / deletion / replacement}: We ask LLM to insert a character to words in the input text at a specified position, or delete a specified character or any character at a specified position from the input text, or replace a character with another character. For example, ``Insert the character \textit{a} to the beginning of all words in the following text: \textit{I am well}" should output ``\textit{aI aam awell}," and similarly for deletion and replacement.

\textbf{Character reordering}:
We provide the LLM with words and ask it to reorder the characters within each word to form a new word, in a similar manner to anagram, e.g., generate ``\textit{epics}" from the input word ``\textit{spice}." The output is deemed correct if it contains all characters in the input word with the same number of occurrences. Note that there is no restriction as to whether new word should be an existing word, as long as all characters have been used. 

\textbf{Character counting}: We provide the LLM with input text and ask it to count the number of certain characters or a category of characters, such as vowels and consonants. For example, ``How many occurrences of the character \textit{s} are in the following word: \textit{obsessed}?" should return 3.

\begin{table*}[t]
\caption{Precision, recall, and F-score for each model on evaluation tasks at character level. For reordering and counting, accuracy is reported in precision column.}
    \label{tab:results}
    \centering
    \scalebox{0.75}{
    \begin{tabular}{c|c|c|c|c|c|c|c|c|c|c|c|c|c|c|c}
        \hline
\multicolumn{1}{c|}{\multirow{2}{*}{Task}} & \multicolumn{3}{c|}{Human} & \multicolumn{3}{c|}{GPT4} & \multicolumn{3}{c|}{Claude} &\multicolumn{3}{c|}{Gemini}&\multicolumn{3}{c}{Mistral}\\
        \cline{2-16}
        & Prec. & Rec. & F-score & Prec. & Rec. & F-score& Prec. & Rec. & F-score & Prec. & Rec. & F-score& Prec. & Rec. & F-score \\
        \hline
        Word Retrieval &\bf 1.0&\bf .989&\bf .994&.523&.691&.595&.406&.534&.461&.549&.602&.574&.614&.671&.641\\
        Character Insertion &\bf 1.0&\bf 1.0&\bf 1.0&.286&.514&.368&.214&.357&.268&.203&.414&.272&.429&.443&.436\\
        Character Deletion &\bf 1.0&\bf 1.0&\bf 1.0&.236&.336&.277&.372&.439&.403&.270&.342&.302&.353&.362&.357\\
        Character Replacement &\bf 1.0&\bf .943&\bf .971&.725&.453&.558&.815&.435&.567&.823&.725&.771&.488&.328&.392\\\hline
        Character Reordering & \bf 1.0&--&--&.91&--&--&.93&--&--&.92&--&--&.88&--&--\\
        Character Counting & \bf .98&--&--&.59&--&--&.51&--&--&.63&--&--&.60&--&--\\
        \hline
    \end{tabular}
    }
    \vspace{-7mm}
\end{table*}

\begin{table}[t]
\caption{F-score for each model on evaluation tasks at token level. For reordering and counting, accuracy is reported.}
    \label{tab:results_token}
    \centering
    \scalebox{0.75}{
    \begin{tabular}{c|c|c|c|c|c}
        \hline
Task & Human & GPT4 & Claude &Gemini&Mistral\\
        \hline
        Sentence Retrieval &\bf 1.0& .926& .893&.921&.953\\
        Word Insertion &\bf 1.0& .625 & .643&.701&.632\\
        Word Deletion & \bf 1.0& .578& .542&.602&.529\\
        Word Replacement & \bf1.0& .991& .994&.993&.981\\\hline
        Word Reordering & \bf .99&.95&.97&.97&.96\\
        Word Counting & \bf 1.0 & .98&.93& .97&.91\\
        \hline
    \end{tabular}
    }
    \vspace{-5mm}
\end{table}

\begin{table*}[t]
\caption{Example failure cases at character level tasks. Bold letters indicate correct answers by the model.}
    \label{tab:failures}
    \centering
    \scalebox{0.9}{
    \begin{tabular}{c|c|c|c|c}
        \hline
        Prompt &  GPT4 & Claude & Gemini & Mistral\\
        \hline
        {\multirow{2}{*}{\shortstack{Find all words with character \textit{o}: \textit{People enjoy music.}\\  (\textit{answer: People, enjoy})}}}&{\multirow{2}{*}{\shortstack{People, enjoy,\\ music}}}&{\multirow{2}{*}{\shortstack{People, enjoy,\\ music}}}&{\multirow{2}{*}{\shortstack{People, enjoy,\\ music}}}&{\multirow{2}{*}{\textbf{People, enjoy}}}\\
        &&&&\\ \hline
        {\multirow{3}{*}{\shortstack{Insert character \textit{i} after the first character for all words:\\ \textit{The Great Wall stretches far.}\\(\textit{answer: Tihe Gireat Wiall sitretches fiar.})}}} &{\multirow{3}{*}{\shortstack{Tihie Giireat\\ Wiiall strietches\\ fari.}}}&{\multirow{3}{*}{\shortstack{iThe iGreat\\ iWall istretches\\ ifar.
}}}&{\multirow{3}{*}{\shortstack{
Thie Great\\ Wall stretches\\ far.}}}&{\multirow{3}{*}{\shortstack{Tihe Griat\\ Wiall striatches\\ fiar.}}}\\ &&&&
        \\&&&&
        \\ \hline
        {\multirow{3}{*}{\shortstack{Remove the third character from the end\\ for all words: \textit{Gravity affects falling objects.}\\(\textit{answer: Gravty affets fallng objets.})}}} &{\multirow{3}{*}{\shortstack{Gravie affect\\ fallin object.}}}&{\multirow{3}{*}{\shortstack{Grav affects\\ fallin objec.
}}}&{\multirow{3}{*}{\shortstack{Gravit affect\\ fallin object.}}}&{\multirow{3}{*}{\shortstack{Gravty affecs\\ faling obects.}}}\\ &&&&
        \\&&&&
        \\ \hline
        {\multirow{2}{*}{\shortstack{Replace all occurrences of \textit{h} with \textit{x}:\\\textit{He has three children.} (\textit{answer: Xe xas txree cxildren.})}}}&{\multirow{2}{*}{\shortstack{xe xas three\\ cxildren.}}}&{\multirow{2}{*}{\shortstack{He xas three\\ children.}}}&{\multirow{2}{*}{\shortstack{\textbf{Xe xas txree}\\ \textbf{cxildren}.}}}&{\multirow{2}{*}{\shortstack{Ex has three\\ children}}}\\&&&&\\ \hline
        {\multirow{3}{*}{\shortstack{Reorder the characters in the following word to form\\ a new word:
        \textit{supercalifragilistic}\\ (\textit{answer: any valid anagram apart from input word itself})}}}&{\multirow{3}{*}{\shortstack{upercalifra-\\gilistic}}}&{\multirow{3}{*}{\shortstack{supercalifr-\\agilistic}}}&{\multirow{3}{*}{\shortstack{lapsticalifr-\\agiceorous}}}&{\multirow{3}{*}{\shortstack{cilisuparegalf-\\itisticxedocious}}}\\
        &&&&\\&&&&\\ \hline
        {\multirow{2}{*}{\shortstack{How many vowels are in the following word:\\ 
        \textit{supercalifragilistic} (\textit{answer: 8})}}}&{\multirow{2}{*}{9}}&{\multirow{2}{*}{11}}&{\multirow{2}{*}{\textbf{8}}}&{\multirow{2}{*}{7}}\\
        &&&&\\ \hline
    \end{tabular}
    }
    \vspace{-5mm}
\end{table*}

\begin{table*}[t]
\caption{Failure cases at token level tasks. Note that they frequently involve numerical elements such as position. Bold letters indicate the correct answer by the model.}
    \label{tab:failures_token}
    \centering
    \scalebox{0.9}{
    \begin{tabular}{c|c|c}
        \hline
        {\multirow{3}{*}{\backslashbox{Model}{Prompt}}} & {\multirow{3}{*}{\shortstack{Remove the third word from the following sentence:\\ \textit{The Renaissance was a period of cultural and artistic rebirth.}\\\textit{(ans: The Renaissance a period of cultural and artistic rebirth.}}}} &  {\multirow{3}{*}{\shortstack{What is the seventh word from the end:\\ \textit{The Great Pyramid of Giza is one of the} \\\textit{Seven Wonders.} (\textit{ans: Giza})}}}\\
        &&\\ &&\\ \hline
        GPT4 & \textit{The was a period of cultural and artistic rebirth.} & \textit{one}\\
        Claude & \textit{The Renaissance was a cultural and artistic rebirth.} &\textit{Wonders} \\
        Gemini & \textit{The Renaissance was a period of artistic rebirth.}&\textit{of}\\
        Mistral & \textbf{\textit{The Renaissance a period of cultural and artistic rebirth.}}&\textit{Pyramid}\\
        \hline        
    \end{tabular}
    }
    \vspace{-5mm}
\end{table*}

We experimented with 4 publicly available LLMs, namely GPT4 \cite{Achiam2023GPT4TR}, Claude \cite{claudeClaude}, Gemini 1.5 \cite{Reid2024Gemini1U}, and Mistral 7B \cite{Jiang2023Mistral7}. We randomly sampled words, phrases, or sentences from Wikipedia corpus. Note that, while it is possible that such publicly available text was used during the pre-training of target LLMs, the character-based nature of our experiments prevents the models from taking advantage of it, and the results in Sec~\ref{sec:results} seem to reinforce the claim. For each task, 100 prompts were used, where each prompt may contain multiple answers. In order to compare the LLM's understanding of character composition with that of humans, we also asked human annotators to perform exactly the same tasks with identical prompts and passages.

In order to compare LLMs' performances at character level and token level tasks, we also extend each task described above to token level tasks. Word retrieval is extended to sentence retrieval, where the model is given 5-sentence passage and is asked to return all sentences containing a target word. Insertion and deletion work similarly by providing target word and position within sentence, whereas we provide target word and another input word for replacement task. Reordering and counting are extended similarly. For reordering, as with character-level reordering, we compute accuracy from whether the final answer is correct, without computing precision and recall for each reordered word.

\subsection{Results}\label{sec:results}

Table~\ref{tab:results} summarizes the results of our experiments with precision, recall, and F-score for each task at character level. For token level, we only report F-score for brevity in Table~\ref{tab:results_token}. It is clearly shown that, for most tasks, all target LLMs display severely degraded performance at character level when compared to token level. While discrepancies exist among respective models' performances, none rises to the level of demonstrating a clear superiority over other models. It is also out of scope of this paper to determine which LLM is better, as our focus is on assessing LLMs in terms of understanding character composition in general. 

Humans, not surprisingly, demonstrated near-perfect performance throughout all tasks. There was hardly any mistake in precision, while defects in recall tended to occur mostly around characters that are placed in the middle of the word, rather than beginning or the end, suggesting attention to saliency in human perception of character composition. Considering that humans have been surpassed by LLMs in many NLP tasks that are supposedly more complex, our results suggest an unsettling dichotomy between LLM's capability at token-level and character-level tasks.

Table~\ref{tab:failures} shows some of the failure cases for each model at character level. It is notable that the tasks for which LLMs struggled the most frequently involved specifying positions of the characters, mostly using numbers, as in insertion or deletion tasks. It should be noted that a similar performance decline was observed even at token level, as illustrated in Table~\ref{tab:results_token}. Table~\ref{tab:failures_token} shows example failure cases at token level. This suggests that some of the limitations in understanding character composition may not simply be attributed to the fact that LLMs are trained at token level, but to a more fundamental drawback in their training approach. Notably, all LLMs performed far better on character reordering task than on other tasks, closely trailing the performance of humans. We conjecture that this is due to abundant resources available online about anagram, which are likely to have been used in pre-training of the models. Even when the newly formed words are non-existing words, many of them are likely to have appeared in the training corpora as possible anagrams of an existing word . It is therefore only natural that all models struggled with character reordering as the word gets longer, or with an unknown word, as shown in Table~\ref{tab:failures_token}.

\begin{table}[t]
\caption{Example of LLMs' performances at token level in tasks that do not involve numerical elements. Bold letters indicate the correct answer by the model.}
    \label{tab:success_token}
    \centering
    \scalebox{0.8}{
    \begin{tabular}{c|c}
        \hline
        {\multirow{3}{*}{\backslashbox{Model}{Prompt}}} & {\multirow{3}{*}{\shortstack{Replace all occurrences of ``the" with ``X":\\ \textit{The history of the city is influenced by the river.}\\(\textit{ans: X history of X city is influenced by X river.})}}} \\
        &\\ &\\ \hline
        GPT4 & \bf \textbf{\textit{X history of X city is influenced by X river.}} \\
        Claude & \bf \textbf{\textit{X history of X city is influenced by X river.}} \\
        Gemini & \textbf{\textit{X history of X city is influenced by X river.}}\\
        Mistral & \textbf{\textit{X history of X city is influenced by X river.}}\\
        \hline        
    \end{tabular}
    }
    \vspace{-7mm}
\end{table}

A clearer contrast between LLMs' performances on token level and character level tasks is made on the tasks that do not involve numerical elements, such as replacement. As illustrated in an example in Table~\ref{tab:success_token}, LLMs rarely have any trouble with replacement task at token level, indicating that token-based embeddings are functioning in a desired manner. Word reordering task also turned out to be reliable, even for fairly long sentences. Such clear contrast between LLMs' performances on token level and character level tasks highlights a fundamental discrepancy in how these models process linguistic information, which suggests that, while LLMs have been effectively optimized for tasks involving tokens, their handling of finer-grained character-level tasks remains inadequate. See Appendix~\ref{appendix} for experiments on languages of varying writing systems.

\section{Discussion}\label{sec:discuss}
As shown throughout the paper, much of limitation in terms of understanding character composition derives from the very nature of LLMs where they are almost invariably trained at token levels, regardless of the pre-training objectives. By operating primarily at the token level, LLMs overlook the intrinsic characteristics and nuances of individual characters within words. This oversight hinders their ability to capture the rich semantic and syntactic information encoded at the character level, leading to sub-optimal performance in tasks requiring fine-grained understanding of language structure. 

A promising direction to address this limitation involves embedding character-level information directly into word embeddings, enabling models to capture the intricate relationships and structures within individual characters. For example, BERT \cite{Devlin2019BERTPO} represents input tokens not only with token embedding, but also with segment embedding, which indicates the sentence that the token belongs to, and position embedding, which shows the position of the token within the sentence. A similar structural approach can be made with respect to character, where character is embedded also with information of the word it belongs to, and its position within the word. Such multi-level embedding strategy could significantly enhance the model's ability to understand and manipulate text at a finer granularity, and can help ensure that the model obtains a robust understanding of word composition while being sensitive to the arrangement of characters within words. Another potential line of approach involves harnessing visual recognition techniques to simulate human-like character perception. In scene text recognition literature, there has been a number of endeavors to integrate computer vision methodologies to visually identify characters, replicating the cognitive processes humans employ when reading and comprehending text \cite{Du2022SVTRST,Bartz2017SEETS}.  By leveraging the complementary strengths of both domains, these approaches may potentially offer novel opportunities for improving robustness for character-level comprehension within large language models. 


\section{Conclusion}
We examined LLMs' ability to understand character composition of words. Our experiments suggest that LLMs utterly fail to demonstrate the ability to understand character composition even at highly simple tasks that can be easily solved by humans with elementary knowledge of language, making a stark contrast with their performances at token level. We further discussed potential future directions, such as incorporating character-embedding and visual features.

\section*{Impact Statement}
This paper presents work whose goal is to advance the field of 
Machine Learning. There are many potential societal consequences 
of our work, none which we feel must be specifically highlighted here.

\nocite{langley00}

\bibliography{example_paper}
\bibliographystyle{icml2024}

\appendix
\section{Appendix}\label{appendix}
Languages of various character systems present vastly different ways of what the characters represent, how syllables are constructed, and how they are tokenized, etc.
We provide an overview of such differences that fundamentally change the way each language is to be processed by language models, focusing on English, Chinese, Korean, and Japanese.

\subsection{Preliminaries}\label{sec:pre}
\begin{CJK}{UTF8}{min}

\begin{table*}[h]
\centering
\caption{Comparison of character systems for English, Chinese, Korean, and Japanese. Space and punctuation marks are skipped in the examples.}
\label{table:compare}
\begin{tabular}{c|c|c|c}
\hline
\textbf{Language} & \textbf{character system} & \textbf{Subword Tokenization} &  \textbf{Example} \\ \hline
\multirow{2}{*}{English} & \multirow{2}{*}{phonetic} & word $\rightarrow$morpheme  & \textit{How are you?} $\rightarrow$ \textit{How}, \textit{are}, \textit{you} \\
&&$\rightarrow$character&   $\rightarrow$ \textit{H,o,w,a,r,e,y,o,u}\\ \hline
Chinese & logographic & word$\rightarrow$character ($\rightarrow$radical)  & \begin{CJK*}{UTF8}{gbsn}你好吗$\rightarrow$你, 好, 吗\end{CJK*} \\ \hline
\multirow{3}{*}{Korean}& \multirow{3}{*}{phonetic, featural} & \multirow{3}{*}{\shortstack{word$\rightarrow$syllable$\rightarrow$letter}} &\begin{CJK}{UTF8}{mj}안녕하세요$\rightarrow$안녕, 하세요\end{CJK} \\ &&& $\rightarrow$\begin{CJK}{UTF8}{mj}안,녕,하,세,요\end{CJK}$\rightarrow$ \\ &&&\begin{CJK}{UTF8}{mj}ㅇ,ㅏ,ㄴ,ㄴ,ㅕ,ㅇ,ㅎ,ㅏ,ㅅ,ㅔ,ㅇ,ㅛ\end{CJK} \\ \hline
\multirow{2}{*}{Japanese} & phonetic,  & word&元気ですか$\rightarrow$元気, です, か\\ &logographic&$\rightarrow$morpheme/character& $\rightarrow$元, 気, で, す, か\\ \hline
\end{tabular}
\vspace{-4mm}
\end{table*}

\begin{table*}[t]
\caption{F-score for each language with different LLMs in evaluation tasks, except accuracy is reported for counting task. For Korean, parentheses indicate the results from letter-based examination, except for insertion and deletion tasks where letter-based examination is not applicable due to structural restraints of Korean characters.}
    \label{tab:results}
    \centering
    \scalebox{0.8}{
    \begin{tabular}{c|*{11}{c|}c}
        \hline
        \multicolumn{1}{c|}{\multirow{2}{*}{Task}} & \multicolumn{3}{c|}{English} & \multicolumn{3}{c|}{Chinese} & \multicolumn{3}{c|}{Korean} &\multicolumn{3}{c}{Japanese}\\
        \cline{2-13}
        & GPT4 & Mist. & Gem. & GPT4 & Mist. & Gem.& GPT4 & Mist. & Gem. & GPT4 & Mist. & Gem. \\
        \hline
        Retrieval & .595 &\bf .641 & .574 &\bf .687 & .535 & .634 & \bf.324 (.229) & .252 (.165)& .279 (.130)&.353 &.328&\bf.407\\
        Insertion & .368 &\bf .436 & .272 & .429 & \bf.437 & .391 & .275 (--)& \textbf{.308} (--) & .201 (--)&.334&.287&\bf.340\\
        Deletion & .277 &\bf .357 &.302 &\bf .387 & .356& .329 & .256 (--)& \textbf{.284} (--)& .239 (--)&\bf.397&.258&.365\\
        Replacement & .558 & .392 &\bf .771& .628 & \bf .743 &.592  &  .361 (\textbf{.110})& \textbf{.485} (.079) & .301 (.058) &.498&.427&\bf.573\\ \hline
        Counting & .59 & .60&\bf .63 & .70&\bf .74 & .65 & \textbf{.49} (\textbf{.27}) & .43 (.20)& .38 (.15)&.55&.50&\bf.57\\
        \hline
    \end{tabular}
    }
    \vspace{-4mm}
\end{table*}

\begin{table*}[t]
\caption{Example failure cases. English translation is provided for each prompt. Bold letters indicate the correct answers from the model. N/A indicates that the model was not able to provide compatible outputs.}
    \label{tab:failures}
    \centering
    \scalebox{0.75}{
    \begin{tabular}{c|c|c|c|c}
        \hline
        Language & Prompt &  GPT4 & Mistral & Gemini \\
        \hline
        {\multirow{3}{*}{English}}&{\multirow{3}{*}{\shortstack{Remove all occurrences of the character \textit{i}\\ Artificial intelligence will revolutionize various fields.\\ (answer: Artfcal ntellgence wll revolutonze varous felds.)}}} &{\multirow{3}{*}{\shortstack{artfcal ntellgence\\ wll revolutonze\\ varous fields.}}}&{\multirow{3}{*}{\shortstack{\bf artfcal ntellgence\\ \bf wll revolutonze\\ \bf varous felds.}}}&{\multirow{3}{*}{\shortstack{artfcial ntellgence\\ wll revolutonze\\ varous felds.}}}\\ &&&& \\ &&&&
        \\ \hline
        {\multirow{3}{*}{Chinese}}&{\multirow{3}{*}{\shortstack{\begin{CJK*}{UTF8}{gbsn}
以下文本中的字符“的”出现了多少次？\end{CJK*}\\ \begin{CJK*}{UTF8}{gbsn}北京的故宫是中国历史文化的重要遗产之一。\end{CJK*}\\Find all characters with letter \begin{CJK*}{UTF8}{gbsn}的\end{CJK*} (answer: 2)}}} &{\multirow{3}{*}{\bf 2}}&{\multirow{3}{*}{4}}&{\multirow{3}{*}{\bf 2}}\\ &&&& \\ &&&&
        \\ \hline
        {\multirow{4}{*}{\shortstack{Korean\\ (character)}}}&{\multirow{3}{*}{\shortstack{\begin{CJK}{UTF8}{mj}
다음 문장에 나오는 문자 ``다"를 모두 ``아"로 바꾸시오:\end{CJK}\\ \begin{CJK}{UTF8}{mj}한국의 전통문화는 다양하고 아름답다.\end{CJK}\\Replace all occurrences of \begin{CJK}{UTF8}{mj}다 with 아\end{CJK} \\(answer:\begin{CJK}{UTF8}{mj} 한국의 전통문화는 아양하고 아름답아.)\end{CJK}}}} &{\multirow{4}{*}{\shortstack{\begin{CJK}{UTF8}{mj}
아아아의\end{CJK}\\ \begin{CJK}{UTF8}{mj}전통문화아는\end{CJK}\\\begin{CJK}{UTF8}{mj}다양하고\end{CJK} \\ \begin{CJK}{UTF8}{mj}아름답다.\end{CJK}}}}&{\multirow{4}{*}{\shortstack{\begin{CJK}{UTF8}{mj}
한국의\end{CJK}\\ \begin{CJK}{UTF8}{mj}전통문화는\end{CJK}\\\begin{CJK}{UTF8}{mj}아다양하고\end{CJK} \\ \begin{CJK}{UTF8}{mj}아름답다.\end{CJK}}}}&{\multirow{4}{*}{N/A}}\\ &&&& \\ &&&& \\ &&&&
        \\ \hline    
        {\multirow{3}{*}{\shortstack{Korean\\ (letter)}}}&{\multirow{3}{*}{\shortstack{\begin{CJK}{UTF8}{mj}
다음 글에서 ㄴ이 들어간 글자를 모두 찾으시오:\end{CJK}\\ \begin{CJK}{UTF8}{mj}안녕하세요 반갑습니다\end{CJK}\\Find all characters with letter \begin{CJK}{UTF8}{mj}ㄴ (answer:안,녕,반,니)\end{CJK}}}} &{\multirow{3}{*}{\begin{CJK}{UTF8}{mj}안,반,갑,습\end{CJK}}}&{\multirow{3}{*}{N/A}}&{\multirow{3}{*}{N/A}}\\ &&&& \\ &&&&
        \\ \hline
        {\multirow{3}{*}{Japanese}}&{\multirow{3}{*}{\shortstack{次の文章で後ろから３番目の文字は何ですか：\\ 吾輩は猫である\\What is the third character from the end (answer:で)}}} &{\multirow{3}{*}{あ}}&{\multirow{3}{*}{\bf で}}&{\multirow{3}{*}{猫}}\\ &&&& \\ &&&&
        \\ \hline
        \end{tabular}
    }
    \vspace{-5mm}
\end{table*}

\textbf{English} employs a phonetic character system based on alphabets, where characters (letters) correspond to sounds that form the basis of words, and multiple characters form a syllable. English words are typically delimited by spaces or punctuation marks, facilitating tokenization at the word level. For example, "\textit{How are you?}" would simply be tokenized into ", ``\textit{how}", ``\textit{are}", ``\textit{you}", ``?". In certain applications, such as parsing, tokens may be further broken down with morphological analysis, \textit{e.g.}, ``\textit{talked}" into ``\textit{talk}" and ``\textit{ed}", but requires additional processes, such as lemmatization or part-of-speech tagging. Splitting English tokens into characters is much more straightforward, as each token corresponds directly to a sequence of characters, such as ``\textit{dog}" being split into ``\textit{d},",``\textit{o},",``\textit{g}."

\textbf{Chinese} utilizes a logographic character system, where characters represent morphemes, words, or meaningful units rather than individual sounds. Each Chinese character carries an inherent meaning and may be composed of various components known as radicals. As such, Chinese characters do not rely on phonetic representation and are not delimited by spaces within words. Chinese tokenization separates text into individual characters or words, depending on the granularity required. For instance, ``\begin{CJK*}{UTF8}{gbsn}老师好，我叫小明。(\textit{Hello teacher, my name is Xiaoming.})" would be tokenized into ``老师(\textit{teacher})", ``好(\textit{hello})", ``，", ``我(\textit{I})", ``叫(\textit{am called})", ``小明(\textit{Xiaoming})", ``。".\end{CJK*} Note that we consider radicals to be out of scope in this paper.

\textbf{Korean} employs a unique featural character system known as Hangul. Korean characters represent both phonetic and featural information, where each character is constructed from combinations of consonants and vowels. For example, consonants \begin{CJK}{UTF8}{mj}ㄱ\end{CJK}(\textit{g}),\begin{CJK}{UTF8}{mj}ㅁ\end{CJK}(\textit{m}) and a vowel \begin{CJK}{UTF8}{mj}ㅣ\end{CJK}(\textit{i}) form a single character \begin{CJK}{UTF8}{mj}김\end{CJK}(\textit{gim}). While structurally combining multiple components to form a single character may seem similar to radicals in Chinese characters, note that each Korean letter represents a phoneme, while radicals are mapped to meanings. As such, any elementary Korean reader can recognize which letters are contained in a character by its sound or its visual composition. Featural structure of Korean poses further challenges for language models as they are encountered not only with character composition of words, but also with how each character is composed. Korean words are typically delimited by spaces or punctuation marks, similarly to English. However, tokenization of Korean can be more complex due to the absence of explicit word delimiters for different parts-of-speech, even within space-delimited chunks. For example, \begin{CJK}{UTF8}{mj}"당신에게(\textit{to you})" involves preposition and pronoun in a single space-delimited chunk, and may be further tokenized into ``당신(\textit{you})", ``에게(\textit{to})"\end{CJK}. It often relies on morphological analysis or dictionary-based approaches to segment text into meaningful units.

\textbf{Japanese} features a complex character system that incorporates elements of both phonetic and logographic scripts. The primary scripts in Japanese are Hiragana and Katakana, which represent phonetic syllables and are used for native Japanese words and loanwords, respectively. Additionally, Japanese utilizes Chinese characters (Kanji) imported from Chinese writing. Kanji characters are logographic and represent morphemes or words, often interspersed with Hiragana or Katakana within sentences. Japanese tokenization involves segmenting text into morphemes, and may utilize dictionary-based methods or rules to identify word boundaries. For example, ``こんにちは、元気ですか？(\textit{Hello, how are you?})" would be tokenized into ``こんにちは(\textit{hello})", ``、", ``元気(\textit{good})", ``です(\textit{is})", ``か(marker for question)", ``？". The mixed nature of Japanese character system poses further challenges for understanding character composition. For example, logographic Kanjis can be converted to phonetic syllables based on their sounds, \textit{e.g.} ``元気" to ``げんき", but it requires the knowledge of the pronunciation, and cannot be determined from the text alone. In this paper, to lower the barrier, we consider Kanjis to be of different characters from their phonetic correspondents.

Table~\ref{table:compare} summarizes the characteristics of each language.

\subsection{Experiments}
\subsubsection{Setting}
We perform the same tasks as in Sec~\ref{sec:experiment} for Chinese, Korean, and Japanese with the exception of word reordering, which is not directly portable over languages of other writing systems. Note that, in many cases, humans will be able to carry out the tasks, even without knowing anything about the target language, via visual inspection. Note that the difficulty posed by each task significantly varies depending on the language. For instance, word retrieval is highly straightforward in Chinese, as words themselves frequently correspond to the target characters. We also use the subset of LLMs used in Sec~\ref{sec:experiment}, namely GPT4, Mistral 7B, and Gemini 1.5. We sampled input texts from Wikipedia of respective languages, and randomly chose a target character that appears at least $N$ times in the sampled text, varying the number from 1 to 3. 100 prompts were used per task for each language. We report F-score for each task in each language. Note that, for Korean, we examined the tasks at two different levels, namely characters and letters (consonants and vowels).

\subsubsection{Results}
Table~\ref{tab:results} summarizes the results across all tasks for each language using different LLMs, and Table~\ref{tab:failures} shows example failure cases for each language. Considering that humans with elementary understanding of respective languages can solve the tasks easily, it is fair to say that all languages perform poorly across all tasks and models. However, there exist notable discrepancies among the performances depending on the language. 

First, LLMs performed slightly better on Chinese than in English throughout the models and tasks, which is notable since, for most models, English is likely the majority language in their training corpora. We conjecture that this may be related to the logographic nature of Chinese character system, where characters are frequently equivalent to words, eliminating the necessity for the models to understand the language at different levels. On the contrary, LLMs' performances unanimously degraded for Korean even at character levels, and utterly failed to demonstrate any sign of understanding the character composition when it comes to further breaking down the characters to letters. In fact, in many cases dealing with Korean letters, LLMs failed even to generate outputs that are compatible for computing accuracy. As discussed in Sec~\ref{sec:pre}, this likely has to with the unique featural structure of Korean language, and suggests that it might require a different approach to account for such irregular systems. Japanese displays similarly lower scores, affirming the complexity of dealing with mixed character system of both logographic and phonetic characters.

\subsubsection{Discussion}
While LLMs' understanding of character composition remains itself an area to be further explored regardless of the language, the results of our experiments seem to further highlight the significant impact of language-specific characteristics on the performance. Multilingual LLMs in general encounter challenges such as uneven distribution of languages in training corpora, or decline in performance for low-resource languages. A popular approach to deal with this problem has been to employ parameter tuning alignment at multiple stages \cite{Qin2024MultilingualLL}. However, it still remains as a challenge to handle language heterogeneity of significantly varying syntax, morphology, and semantics \cite{Xu2024ASO}, and our results seem to further reinforce the claim that there may be an underlying limitation to conventional approaches, without accounting for fundamental differences between the language systems.

\end{CJK}
\end{document}